\documentclass[10pt, a4paper]{article}
\usepackage[]{lrec-coling2024} 

\usepackage{booktabs}
\usepackage{multirow}
\usepackage{graphicx}
\usepackage{subcaption}
\usepackage{natbib}
\usepackage{multibib}
\usepackage{colortbl}
\usepackage{kotex}
\usepackage{placeins}
\usepackage{footnote}
\usepackage{mdframed}
\usepackage{xstring}
\usepackage{color}
\usepackage{xcolor}
\usepackage{hyperref}
\usepackage{amsmath}
\usepackage{amssymb}
\mdfsetup{
    linecolor=white,
    backgroundcolor=gray!20,
}
\definecolor{darkblue}{rgb}{0, 0, 0.5}
\hypersetup{colorlinks=true, citecolor=darkblue, linkcolor=darkblue, urlcolor=darkblue}

\title{CLIcK: A Benchmark Dataset of Cultural and Linguistic Intelligence in Korean}

\name{Eunsu Kim$^{\diamond}$, Juyoung Suk$^{\dagger}$, Philhoon Oh$^{\dagger}$, Haneul Yoo$^{\diamond}$, James Thorne$^{\dagger}$, Alice Oh$^{\diamond}$} 

\address{$^{\diamond}$School of Computing, KAIST, Dajeon, South Korea\\
        $^{\dagger}$GSAI, KAIST, Seoul, South Korea \\
        \{kes0317, scottsuk0306, philhoonoh, haneul.yoo, thorne\}@kaist.ac.kr , alice.oh@kaist.edu}

\date{}

\abstract{Despite the rapid development of large language models (LLMs) for the Korean language, there remains an obvious lack of benchmark datasets that test the requisite Korean cultural and linguistic knowledge. Because many existing Korean benchmark datasets are derived from the English counterparts through translation, they often overlook the different cultural contexts. For the few benchmark datasets that are sourced from Korean data capturing cultural knowledge, only narrow tasks such as bias and hate speech detection are offered. To address this gap, we introduce a benchmark of Cultural and Linguistic Intelligence in Korean (CLIcK), a dataset comprising 1,995 QA pairs. CLIcK sources its data from official Korean exams and textbooks, partitioning the questions into eleven categories under the two main categories of language and culture. For each instance in CLIcK, we provide fine-grained annotation of which cultural and linguistic knowledge is required to answer the question correctly. Using CLIcK, we test 13 language models to assess their performance. Our evaluation uncovers insights into their performances across the categories, as well as the diverse factors affecting their comprehension. CLIcK offers the first large-scale comprehensive Korean-centric analysis of LLMs' proficiency in Korean culture and language. CLIcK is publicly available at: \url{https://github.com/rladmstn1714/CLIcK }.
 \\ \newline }
\begin{document}

\maketitleabstract

\section{Introduction}

Recent advancements in Large Language Models (LLMs) have been significant, particularly for a small group of high-resource languages including English. In these languages, LLMs frequently attain or surpass human-level proficiency in numerous Natural Language Processing (NLP) tasks that necessitate comprehension of everyday life and the subtleties of linguistic nuances.
However, despite a concerted effort in developing Korean large-scale language models, there remains a significant performance gap for benchmark tasks in the Korean language.
For example, KoGPT3-39B underperforms on the Korean HellaSwag task~\citelanguageresource{jang-etal-2022-kobest} by 20\%, compared to similar scale English Falcon40B model~\cite{almazrouei2023falcon} in the original English version of the task~\citelanguageresource{zellers-etal-2019-hellaswag} even though human annotators can attain the same performance.
Instances that contain cultural and linguistic knowledge that deviate from English and other well-represented languages are often incorrectly answered by models. 
~\begin{figure}
    \centering\includegraphics[width=0.9\columnwidth]{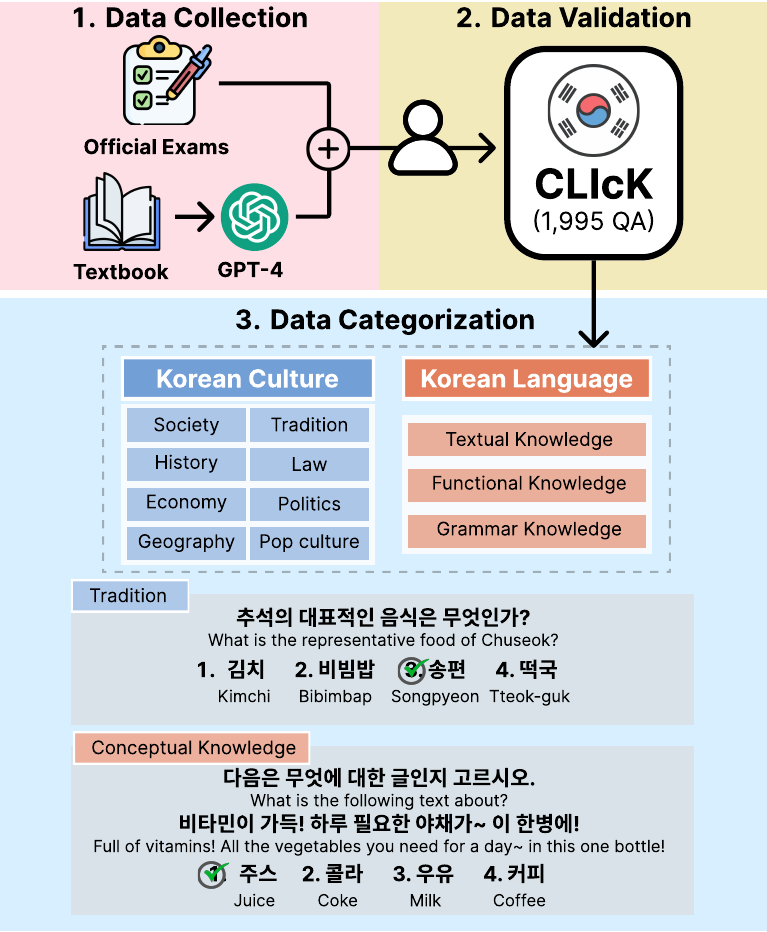}
    \caption{Overview of the CLIcK dataset curation and categorization process. Data is sourced from official exams and textbooks and validated by authors. The dataset is categorized into Korean Culture and Korean Language, further divided into 11 sub-categories.}
    \label{fig:teaser_image}
\end{figure}

Current Korean evaluation datasets for LLMs show significant limitations for comprehensive assessment. Existing tasks are either too simple~\citelanguageresource{ham-etal-2020-kornli, park2021klue} or mainly derived from English benchmarks\thinspace\footnote{\url{https://huggingface.co/spaces/upstage/open-ko-llm-leaderboard}\label{openkollm}}, failing to capture the key aspects of the Korean language or culture. While several Korean-centric datasets have been introduced, these typically target specific tasks such as bias and hate speech detection~\citelanguageresource{jin2023kobbq,jeong-etal-2022-kold} and hence cannot support general LLM evaluation.

To bridge this resource gap, we construct and release CLIcK, a culturally-aware evaluation benchmark dataset encompassing 1,995 instances across 11 categories representing facets of the Korean culture, ranging from everyday life to specific subject areas, as well as Korean grammar and linguistics.
To ensure high quality, we collect and curate all samples from official examinations and textbooks, which are then rigorously inspected and categorized by four native speakers of Korean.

We subsequently evaluate five families of open-source LLMs with different parameter sizes and two proprietary LLMs with CLIcK. The open-source models exhibit especially low accuracies, ranging between 10\% and 50\%. Meanwhile, proprietary LLMs like \texttt{GPT-3.5} and \texttt{Claude-2} outperform these but still perform poorly in some categories.
Notably, compared to the general population of Korean test-takers, \texttt{GPT-3.5} scores in the lowest 11th percentile, which contrasts with its achievement in the top 13th percentile on the English Scholastic Aptitude Test (SAT).

The primary contributions of our work include:
\begin{itemize}
    \item We construct and publicly release CLIcK, a benchmark dataset to evaluate LLMs' cultural and linguistic understanding of Korean.
    \item We provide a fine-grained categorization of the requisite knowledge to answer each query in the dataset.
    \item We empirically evaluate 13 model configurations on CLIcK, demonstrating the limitations of LLMs and motivating further research on cultural and linguistic benchmarks.
\end{itemize}

\section{Related Work}
\subsection{General English Benchmarks}
The General Language Understanding Evaluation (GLUE) and SuperGLUE benchmarks were established to evaluate models in a wide range of tasks such as sentiment analysis, natural language inference, and question-answering~\citelanguageresource{wang-etal-2018-glue,wang2020superglue}. HellaSwag~\citelanguageresource{zellers-etal-2019-hellaswag} and CosmoQA~\citelanguageresource{huang-etal-2019-cosmos} further include commonsense reasoning for a more robust evaluation. However, due to the rapid advancement in AI communities, achieving human-level state-of-the-art performances on these datasets has become increasingly common~\cite{Martinez-Plumed2021}. To properly evaluate the capabilities of models in the era of LLMs, more challenging benchmarks have been introduced. For example, MMLU~\citelanguageresource{hendryckstest2021} and AGIEval~\citelanguageresource{zhong2023agieval} consist of questions originally designed for humans, while BIG-bench~\citelanguageresource{srivastava2023imitation} aims to cover diverse topics, comprising 204 tasks to address the limitations in LLMs.


\subsection{Multilingual and Commonsense}

To investigate how LMs can comprehend or generate text in other languages, there have been efforts to construct multilingual datasets. For instance, XGLUE~\citelanguageresource{liang-etal-2020-xglue} encompasses 100 languages that can be employed for both pre-training and evaluating cross-lingual tasks. XTREME~\citelanguageresource{hu2020xtreme} introduces an evaluation framework for cross-lingual benchmarks, while MEGA~\citelanguageresource{ahuja2023mega} focuses on assessing LLMs, providing 16 NLP datasets ranging from low-resource to high-resource languages. In addition, datasets that primarily focus on specific target languages, such as Chinese, Indian, and African languages, have been introduced~\citelanguageresource{huang2023ceval, doddapaneni2023leaving, adebara2023serengeti}.

The popularity of commonsense datasets has increased because they reflect a wide array of sociocultural knowledge shared by humans~\cite{commonsense-article}. These datasets incorporate everyday concepts, such as CommonsenseQA~\citelanguageresource{talmor-etal-2019-commonsenseqa}, scientific knowledge like ARC~\citelanguageresource{clark2018think}, and simple arithmetic reasoning like GSM8K\citelanguageresource{cobbe2021gsm8k}. These datasets can be seen as a representation of general and practical knowledge that aligns with human intentions. Consequently, certain datasets incorporate or employ translated portions from English datasets~\citelanguageresource{seo-etal-2022-dog}, potentially overlooking subtle linguistic or cultural differences that may not be apparent to the audience~\cite{10.1145/3186549.3186562}.

~\citealt{lee-etal-2023-hate} demonstrated that language models fail to capture biases in different languages due to their cultural insensitivity, which can have societal impacts ~\cite{tamkin2021understanding}. Furthermore, \citealt{ma-etal-2022-encbp} emphasized the importance of cultural background and showed that integrating cultural knowledge can improve models performance. These findings illustrate the need for cultural evaluation datasets. However, building a cultural evaluation dataset from scratch is challenging since it entails significant time and resources while relying on translated datasets fails to incorporate cultural knowledge in different languages.

\subsection{Korean Datasets}
Previous benchmarks for Korean language models focused on specific tasks such as paraphrase detection, natural language inference (NLI), machine reading comprehension (MRC), and hate speech detection~\citelanguageresource{yang-etal-2019-paws, ham2020kornli, lim2019korquad10, moon-etal-2020-beep}. The Korean Language Understanding Evaluation (KLUE) benchmark~\citeplanguageresource{park2021klue}, similar to GLUE, introduced eight downstream tasks for the Korean language. However, these benchmarks lacked tasks for advanced reasoning, which is inadequate for LLM evaluation. Recently published Open Ko-LLM LeaderBoard\thinspace\textsuperscript{\ref{openkollm}} aimed to address this issue; however, its closed-source nature and reliance on translations may not fully represent the nuances of the Korean language context.

Contemporary benchmarks prioritize preserving linguistic and cultural nuances of the target language in translation. For instance, \citetlanguageresource{jin2023kobbq} introduced the Korean Bias Benchmark for Question Answering (KoBBQ) based on the original BBQ dataset~\citeplanguageresource{parrish-etal-2022-bbq}. KoBBQ first classified the translations into 3 distinct categories: \textsc{Simply-Translated}, where translations are appropriate for the Korean knowledge, \textsc{Sample-Removed} where translated sentences are removed due to their lack of relevance to Korean cultural context, and \textsc{Target-Modified} where target translations are adjusted to align with Korean culture background. This classification adjusts knowledge differences available in languages and reflects subtle cultural nuances in Korean.

Recent efforts have revolved around creating authentic Korean datasets from scratch. For instance, the Korean Offensive Language Dataset~\citelanguageresource{jeong-etal-2022-kold} gathers toxic sentences from news articles and YouTube platforms, while the Korean Balanced Evaluation of Significant Tasks~\citelanguageresource{jang-etal-2022-kobest} is entirely annotated by humans for five distinct tasks. The HAE-RAE Benchmark~\citelanguageresource{son2023haerae} provides Korean Reading Comprehension datasets sourced from the original Korean Corpus. However, these datasets are constrained to specific tasks and may not be suitable for evaluating diverse topics related to Korean cultural and linguistic knowledge. With the rapid growth of the language model's capacity, there is a growing need for fine-grained evaluation to assess the cultural and commonsense knowledge within LLMs~\cite{ye2023flask}.


In this paper, we introduce CLIcK comprising 1,995 questions that require a wide range of Korean linguistic and cultural knowledge along with reasoning capacity, which is close to real-world settings. Additionally, it directly sources content from the original Korean Corpus, including six different Korean native exams, ensuring cultural authenticity and facilitating comparisons between models and human-level scores. Lastly, CLIcK provides fine-grained evaluations of LLMs on eleven diverse topics, thereby contributing to further research on the assessment of Korean cultural and linguistic knowledge within LLMs.

\section{CLIcK Dataset}
CLIcK contains 1,995 QA pairs, organized into two main categories and 11 subcategories of multiple-choice QA about Korean facts. Dataset statistics are reported in Tables~\ref{tab:data_stat} and~\ref{tab:stats_exam}.
The dataset is constructed in three stages: (1) Data Collection(\S~\ref{sec:data_collection}), (2) Data Validation(\S~\ref{sec:data_valid}), and (3) Data Categorization(\S~\ref{sec:data_cat}), summarized in Figure~\ref{fig:teaser_image}.
~\begin{table}[htb!]
\resizebox{\columnwidth}{!}{
\begin{tabular}{@{}ll|rrr@{}}
\toprule
\multicolumn{2}{c|}{\multirow{2}{*}{Category}}                                                                & \multicolumn{3}{c}{\# of Samples}                                                                    \\ \cmidrule(l){3-5} 
\multicolumn{2}{c|}{}                                                                           & \multicolumn{1}{l|}{Textbook} & \multicolumn{1}{l|}{Exams} & \multicolumn{1}{l}{Total} \\ \midrule
\multicolumn{1}{l|}{\multirow{8}{*}{\begin{tabular}[c]{@{}l@{}}Korean\\ Culture\end{tabular}}}  & Society     & \multicolumn{1}{r|}{284}              & \multicolumn{1}{r|}{25}            & 309                         \\
\multicolumn{1}{l|}{}                                                                           & Tradition   & \multicolumn{1}{r|}{161}              & \multicolumn{1}{r|}{61}            & 222                         \\
\multicolumn{1}{l|}{}                                                                           & History     & \multicolumn{1}{r|}{0}              & \multicolumn{1}{r|}{280}            & 280                         \\
\multicolumn{1}{l|}{}                                                                           & Law         & \multicolumn{1}{r|}{51}              & \multicolumn{1}{r|}{168}            & 219                         \\
\multicolumn{1}{l|}{}                                                                           & Politics    & \multicolumn{1}{r|}{79}              & \multicolumn{1}{r|}{5}            & 84                         \\
\multicolumn{1}{l|}{}                                                                           & Economy     & \multicolumn{1}{r|}{57}              & \multicolumn{1}{r|}{2}            & 59                         \\
\multicolumn{1}{l|}{}                                                                           & Geography   & \multicolumn{1}{r|}{39}              & \multicolumn{1}{r|}{92}            &  131                        \\
\multicolumn{1}{l|}{}                                                                           & Pop culture & \multicolumn{1}{r|}{15}              & \multicolumn{1}{r|}{26}            & 41                         \\ \midrule
\multicolumn{1}{l|}{\multirow{3}{*}{\begin{tabular}[c]{@{}l@{}}Korean\\ Language\end{tabular}}} & Textual  & \multicolumn{1}{r|}{0}              & \multicolumn{1}{r|}{285}            & 285                         \\
\multicolumn{1}{l|}{}                                                                           & Functional  & \multicolumn{1}{r|}{0}              & \multicolumn{1}{r|}{133}            & 133                         \\
\multicolumn{1}{l|}{}                                                                           & Grammar     & \multicolumn{1}{r|}{0}              & \multicolumn{1}{r|}{232}            & 232                         \\ \midrule
\multicolumn{1}{l}{}                                                                           & Total     & \multicolumn{1}{r|}{1245}              & \multicolumn{1}{r|}{750}            & 1995                         \\ \bottomrule
\end{tabular}
}
\caption{Statistics of CLIcK per categories. ‘From Textbook' denotes data from the KIIP textbook; ‘From Exams' refers to data from official exams. ‘Number of samples' indicates unique QA pairs in the dataset.}
\label{tab:data_stat}
\end{table}
~

\begin{table}[htb!]
\centering
\resizebox{0.65\columnwidth}{!}{
\begin{tabular}{@{}ll|r@{}}
\toprule
Code  & Subject      & \# of Samples\\ 
\midrule
CSAT  & Language     & 226         \\
      & Geography    & 30          \\ 
\midrule
TOPIK & Language     & 237         \\ 
\midrule
PSE   & Language     & 14          \\
      & History      & 189         \\ 
\midrule
PSAT  & Constitution & 168         \\ 
\midrule
KHB   & History      & 47          \\ 
\midrule
Kedu  & Language     & 173         \\
      & Culture      & 161         \\ 
\bottomrule
\end{tabular}
}
\caption{Statistics of CLIcK per Exams. Code and Subject refer to the exam's codes and its utilized subjects.}
\label{tab:stats_exam}
\end{table}

\subsection{Data Collection}
\label{sec:data_collection}
To collect data, we employ two approaches: 1) following the AGIEval dataset~\citelanguageresource{zhong2023agieval} to select questions from standardized Korean exams, and 2) using GPT-4 to generate new questions based on textbooks.
In this process, we utilize six exams and one textbook related to Korean language and cultural knowledge as sources (the genres of these resources are summarized in Table~\ref{tab:source}).

\paragraph{Selecting Questions from Exams}
We obtain test data from six Korean examinations and receive permission from the relevant institutions. Descriptions for each examination can be found in Appendix~\ref{sec:appendix_A}.
We use Clova OCR\footnote{\url{https://clova.ai/ocr/}} to extract text from the exams, excluding images and tables.
\paragraph{Generating Questions Using GPT-4} We use this approach to introduce novel cultural questions, which are not covered in the exams. Building on established practices in question generation~\citep{Zhou2017NeuralQG, Kurdi2019ASR}, we feed \texttt{GPT-4} the full text of each chapter from the KIIP textbook, prompting it to produce multiple-choice questions, their corresponding choices and answers based strictly on the book's content. Each question was verified before inclusion in the dataset.
More detailed information on GPT-4 question generation, including used prompts and procedures, can be found in Appendix~\ref{sec:appendix_B}.

\begin{table}[t]
\resizebox{\columnwidth}{!}{%
\begin{tabular}{@{}l|l|l|l@{}}
\toprule
Source                                                                                & Code  & Type     & Subject         \\ \midrule
\begin{tabular}[c]{@{}l@{}}Korean Immigration \\ and Integration Program\end{tabular} & KIIP  & Textbook & Culture  \\ \midrule
\begin{tabular}[c]{@{}l@{}}College Scholastic \\ Ability Test of Korea\end{tabular} &
  CSAT &
  Exam &
  \begin{tabular}[c]{@{}l@{}}Language\\ Geography\end{tabular} \\ \midrule
\begin{tabular}[c]{@{}l@{}}Test of Proficiency \\ in Korean\end{tabular}              & TOPIK & Exam     & Language \\ \midrule
\begin{tabular}[c]{@{}l@{}}National Public Service \\ Examination - Grade 9\end{tabular} &
  PSE &
  Exam &
  \begin{tabular}[c]{@{}l@{}}Language\\ History\end{tabular} \\ \midrule
Public Service Aptitude Test                                                            & PSAT   & Exam     & Constitution  \\ \midrule
Korean History Exam-Basic                                                             & KHB   & Exam     & History  \\ \midrule
\begin{tabular}[c]{@{}l@{}}Test of Teaching Korean \\ as a Foreign Language\end{tabular} &
  Kedu &
  Exam &
  \begin{tabular}[c]{@{}l@{}}Language\\ Culture\end{tabular} \\ \bottomrule
\end{tabular}%
}
\caption{Overview of data sources used in the CLIcK dataset, detailing each source’s code, type (textbook or exam), and covered subjects related to Korean language and culture.}
\label{tab:source}
\end{table}
\subsection{Data Validation}
\label{sec:data_valid}
We scrutinize transcribed exams for OCR errors and validate the \texttt{GPT-4} generated questions according to the following criteria:
\begin{mdframed}
\scriptsize{1. Questions are solely based on the given text.\\
2. Information in Questions remains consistent over time.\\
3. Questions should centrally relate to Korea. \\
4. Questions should be objective and free from bias. 


}
\end{mdframed}
By checking the first criterion, we ensure that questions originate exclusively from the textbook's content, eliminating any influence of \texttt{GPT-4}'s internal knowledge. Additionally, the fourth criterion helps maintain dataset objectivity by excluding instances connected to subjective beliefs or biases, like gender, politics, or international relations.

\paragraph{Human Annotation}
Four of the authors, who are Korean native speakers, validated to ensure its relevance and quality. Validation checks are categorized into three types; valid, needs modification, and invalid, with the following process:
\begin{enumerate}
    \item Initially, three annotators reviewed each sample. If two or more annotators considered a sample invalid, it was discarded. 15.9\% of the data was labeled invalid by one annotator. Samples that needed modification were revised by one of the annotators.
    \item For the remaining invalid and modified samples, a second round of annotation was conducted by three annotators. After this phase, 3.9\% of the initial set still had discrepancies.
    \item The four annotators involved in the previous steps discuss the disagreements. Only samples with unanimous agreement between all four annotators are included in the dataset.
\end{enumerate}
Following the validation process, the initial dataset, which consisted of 1,985 samples, was reduced to 1,245 samples, accounting for 62.9\% of the original dataset.
\subsection{Data Categorization}
\label{sec:data_cat}
For each instance, we provide fine-grained annotation of which aspects of Cultural and Linguistic intelligence are required to answer the question. Summary statistics are presented in Table~\ref{tab:data_stat}.
\paragraph{Cultural Intelligence} We adopt eight subcategories based on the KIIP textbook. 
The primary chapters of the KIIP basic textbook encompass \textit{Society, Culture, Politics, Economy, Law, History,} and \textit{Geography}. Within the \textit{Culture} chapter, there are subsections on \textit{Tradition} and \textit{Pop Culture}. We label each instance with respect to its cultural knowledge with one of the following labels: \textit{Society, Tradition, Pop Culture, Politics, Economy, Law, History}, and \textit{Geography}.
\paragraph{Linguistic Intelligence} 
We follow definitions of linguistic knowledge from ~\citet{bachman1996language}. Specifically, we annotate instances for \textit{Textual Knowledge}, which concerns organizing utterances into coherent texts with cohesion and rhetorical structures; \textit{Functional Knowledge}, focusing on the communicative roles of language, especially ideational, manipulative, heuristic, and imaginative functions; and \textit{Grammatical Knowledge}, addressing the organization of utterances with an emphasis on vocabulary, syntax, and phonology/graphology. We exclude the \textit{socio-linguistic knowledge} category in our work because it is largely subsumed by our annotations in the Cultural Intelligence category.
\paragraph{Specific Procedures}
Because questions generated from the textbook already align to Cultural Intelligence categories, we do not require further annotation for these instances. Similarly, for exams that focus on a single subject (e.g. \textit{History}), no additional categorization is performed. 
For the CSAT-Korean exam, problems are categorized into speaking, writing, language, media, literature, and reading. In terms of our defined categories, speaking and writing correspond to \textit{Functional knowledge}, language aligns with \textit{Grammar knowledge}, and both literature and reading are associated with \textit{Textual knowledge}. Other Korean Language exams offer solutions detailing the problem's category. Based on this information, a single annotator validates the label assignment.
\section{Evaluation}
We evaluate the capabilities of established language models using our CLIcK dataset. We compare a range of models which have a variety of exposure to the Korean language (detailed in Table~\ref{tab:model}). For API-based LLMs, we conduct experiments between September and October of 2023.
\paragraph{Prompt Types}\label{prompt}\ Our prompts are derived from ~\citetlanguageresource{jin2023kobbq}. Following ~\citet{izacard_few-shot_2022}, we apply cyclic permutation for each question to mitigate option's order effects in the prompt to the language model. We report the average over three different wordings of the prompt.  Depending on whether the question requires the model to read background information, we prompt the model with context (type 1) or without context (type 2) according to the examples which are provided below.
~
\begin{table}[t!]
\resizebox{\columnwidth}{!}{
\begin{tabular}{@{}l|ll@{}}
\toprule
Type                                                                       & \multicolumn{2}{l}{Model}                                                                                                                                                                                 \\ \midrule
\begin{tabular}[c]{@{}l@{}}API-based\\ LLMs\end{tabular}                 & \multicolumn{2}{l}{\begin{tabular}[c]{@{}l@{}}GPT-3.5-turbo, Claude-2\end{tabular}}                                                                                                                      \\ \midrule
\multirow{2}{*}{\begin{tabular}[c]{@{}l@{}}Open-source\\ LMs\end{tabular}} & \multicolumn{1}{l|}{Multilingual}                                                   & Korean-specialized                                                                                                 \\ \cmidrule(l){2-3} 
& \multicolumn{1}{l|}{\begin{tabular}[c]{@{}l@{}}LLaMA2-chat\\ (7B,13B)\end{tabular}} & \begin{tabular}[c]{@{}l@{}}LLaMA2-Ko(7B)\hyperlink{fn:llama-ko}{\footnotemark}\\ KULLM-v2~\citep{lee2023kullm}\\ KoAlpaca\hyperlink{fn:koalpaca}{\footnotemark}\\ Polyglot-Ko~\citep{ko2023polyglotko} \\ (1.3B,3.8B,5.8B,12.8B)\end{tabular} \\ \bottomrule
\end{tabular}
}
\caption{Model Selection for Our Experiment. The numbers in parentheses indicate the parameters for models.}
\label{tab:model}
\end{table}

\addtocounter{footnote}{-1} 
\footnotetext{\hypertarget{fn:llama-ko}{}\url{https://huggingface.co/beomi/llama-2-ko-7b}}
\stepcounter{footnote}
\footnotetext{\hypertarget{fn:koalpaca}{}\url{https://github.com/Beomi/KoAlpaca}}

\begin{mdframed}
\scriptsize{\textbf{Type 1}: Sample with Context\\주어진 맥락을 천천히 읽고, 질문에 대한 적절한 정답을 A, B, C, D 중에 골라 알파벳 하나로 답하시오.\\
(Read the given context, and choose the correct answer to the question from options A, B, C, or D. Respond with a single alphabet.)\\
\\
맥락 (Context): \{CONTEXT\}\\
질문 (Question): \{QUESTION\}\\
보기 (Options):\\
A: \{A\}, 
B: \{B\}, 
C: \{C\}, 
D: \{D\}\\
정답 (Answer):
}
\end{mdframed}
\begin{mdframed}
\scriptsize{\textbf{Type 2}: Sample without Context\\주어진 질문을 천천히 읽고, 적절한 정답을 A, B, C, D 중에 골라 알파벳 하나로 답하시오.\\
(Read the given Question, and choose the correct answer from options A, B, C, or D. Respond with a single alphabet.)\\
\\
질문 (Question): \{QUESTION\}\\
보기 (Options):\\
A: \{A\}, 
B: \{B\}, 
C: \{C\}, 
D: \{D\}\\
정답 (Answer):
}
\end{mdframed}
\paragraph{Evaluation Methodology} We adopt the evaluation methodology from MMLU, which also aligns with prevalent LLM evaluation frameworks such as EleutherAI lm-harness\thinspace\footnote{\url{https://github.com/EleutherAI/lm-evaluation-harness}} and OpenAI Evals\thinspace\footnote{\url{https://github.com/openai/evals}}. For open-source models, we examine the output probabilities of option ID tokens (A/B/C/D or A/B/C/D/E) concatenated with the option string, selecting the most probable answer as the model prediction.
For API-based LLMs (\texttt{GPT-3.5-turbo} and \texttt{Claude-2}), the evaluation involves comparing the generated response with the labeled answer. Here, the decoding temperature is set to 0. Though our prompt directly asks the model to output only the option ID, we notice that these models may at times produce verbose responses. Therefore, we adopt the acceptance criteria from \citetlanguageresource{jin2023kobbq}, allowing answers that: i) mention only one alphabet from the given options, ii) exactly match a term provided in the options, iii) include specific expressions clearly intended to convey the answer, such as ‘the answer is -’, or iv) present the answer distinctly as per conditions i) to iii), followed by further explanation. Responses that don’t meet these conditions are considered as out-of-option answers.
~\begin{table*}[tb!]
\resizebox{\textwidth}{!}{%
\begin{tabular}{@{}cc|cccc|cc|cc|c|cc|c|c@{}}
\toprule
\multicolumn{2}{c|}{\multirow{2}{*}{}}                          
& \multicolumn{4}{c|}{\texttt{Polyglot-Ko}}                              
& \multicolumn{2}{c|}{\texttt{KULLM}}
& \multicolumn{2}{c|}{\texttt{KoAlpaca}}
& \multicolumn{1}{c|}{\texttt{LLaMA-Ko}}
& \multicolumn{2}{c|}{\texttt{LLaMA}}
& \multirow{2}{*}{\texttt{GPT-3.5} }
& \multirow{2}{*}{\texttt{Claude2}} \\ \cmidrule(lr){3-13}
\multicolumn{2}{c|}{}                                           
& 1.3B 
& 3.8B
& 5.8B
& 12.8B
& 5.8B
& 12.8B
& 5.8B
& 12.8B
& 7B
& 7B
& 13B
& 
&                         \\ \midrule
\multicolumn{1}{c|}{\multirow{9}{*}{\begin{tabular}[c]{@{}c@{}}Korean\\ Culture\end{tabular}}} 
& History & 26.30 & 24.71 & 25.52 & 24.43 & 26.48 & 25.07 & 26.05 & 25.84 & 26.38 & \textcolor{blue}{30.75} & 30.73 & 31.32 & \textbf{35.00} \\
\multicolumn{1}{c|}{}                                                                           & Geography & 30.18 & 28.72 & 29.06 & 30.12 & 27.21 & 28.66 & 28.53 & 30.01 & \textcolor{blue}{33.21} & 23.10 & 25.20 & \textbf{45.42} & 43.30 \\
\multicolumn{1}{c|}{}                                                                           & Law & 38.44 & 40.16 & 40.70 & 43.44 & 41.67 & 41.90 & 40.67 & 40.13 & 40.02 & \textcolor{blue}{45.13} & 44.12 & 55.31 & \textbf{57.09} \\
\multicolumn{1}{c|}{}                                                                           & Politics & 30.53 & 32.00 & 27.74 & 27.15 & 26.96 & 22.68 & 23.42 & 28.79 & \textcolor{blue}{36.03} & 27.31 & 26.43 & 47.75 & \textbf{60.89} \\
\multicolumn{1}{c|}{}   & Society & 34.69 & 34.31 & 35.95 & 37.37 & 35.95 & 37.37 & 33.33 & 36.44 & 32.10 & 39.48 & \textcolor{blue}{40.93} & 60.48 & \textbf{62.43} \\
\multicolumn{1}{c|}{}                                                                           & Tradition & 32.48 & 34.01 & 34.97 & 33.96 & 35.86 & 34.63 & 32.80 & 35.45 & 33.60 & 33.88 & \textcolor{blue}{36.12} & 50.16 & \textbf{52.10} \\
                                                            
\multicolumn{1}{c|}{}                                                                           & Economy & 42.54 & 42.62 & 42.25 & 45.03 & 43.86 & 44.08 & 43.35 & 43.79 & 45.32 & 45.83 & \textcolor{blue}{46.27} & 47.59 & \textbf{53.62} \\

\multicolumn{1}{c|}{}                                                                           & Pop culture & 29.77 & 33.68 & 32.64 & 29.59 & 33.60 & 32.76 & 34.02 & 32.63 & 27.20 & 33.45 & \textcolor{blue}{36.41} & \textbf{68.61} & 59.56 \\ \cmidrule{2-15}
\multicolumn{1}{c|}{}                                                                           & Average & 32.71 & 32.90 & 33.14 & 33.40 & 33.79 & 33.51 & 32.33 & 33.80 & 33.26 & 35.44 & \textcolor{blue}{36.22} & 49.30 & \textbf{51.72} \\
\midrule
\multicolumn{1}{c|}{\multirow{4}{*}{\begin{tabular}[c]{@{}c@{}}Korean\\ Language\end{tabular}}} & Textual & 23.44 & 23.57 & 23.27 & 22.96 & 24.52 & 24.65 & 23.07 & 24.19 & \textcolor{blue}{26.75} & 24.73 & 24.29 & 53.19 & \textbf{55.86 }\\
\multicolumn{1}{c|}{}                                                                           & Functional & 23.77 & 21.67 & 22.64 & 19.84 & 20.06 & 19.38 & 24.76 & 20.50 & 26.31 & 27.04 & \textcolor{blue}{30.50} & 32.62 & \textbf{32.88}\\
\multicolumn{1}{c|}{}                                                                           & Grammar & 21.87 & 21.79 & 23.64 & 23.05 & 24.69 & 25.67 & 24.03 & 22.05 & 23.04 & \textcolor{blue}{29.32} & 26.52 & 38.85 & \textbf{43.95} \\ \cmidrule{2-15}
\multicolumn{1}{c|}{}                                                                           & Average & 22.88 & 22.38 & 23.27 & 22.24 & 23.50 & 23.78 & 23.87 & 22.42 & 25.69 & \textcolor{blue}{27.17} & 26.71 & 42.32 & \textbf{45.39} \\ \bottomrule 
\end{tabular}
}
\caption{Accuracy of the models by category. The highest accuracy for each category is in bold. The top-performing open-source models are marked in blue.}
\label{Tab:acc_results}
\end{table*}

\paragraph{Evaluation Metric}
Our primary evaluation metric is accuracy. We report the average accuracy over the entire dataset. As we prompt the model 3 times and adopt cyclic permutation for each instance, the total number of experiments per instance is $3N$, where $N$ denotes the ``number of options''. Instance accuracy is computed as:
\begin{equation}
p_{\text{accuracy}} = \frac{\text{count}(\text{correct answers})}{3N}
\end{equation}
\subsection{Experimental Results}
We present comprehensive results of evaluating 13 LLMs with CLIcK in Table~\ref{Tab:acc_results}.
We find that open-source models with fewer than 13B parameters generally exhibit a low accuracy in the range of 10-50\%.
While \texttt{Claude-2} and \texttt{GPT-3.5} surpass those smaller models in most categories, their performance remains similar in \textit{History, Economy}, and \textit{Functional Knowledge}. \texttt{Claude-2} outperforms \texttt{GPT-3.5} in all categories except \textit{Geography} and \textit{Pop Culture}.

~
\begin{table}[t!]
\centering
\small{
\begin{tabular}{@{}l|ll@{}}

\toprule
Factor                                                & \textit{F} & \textit{p} \\ \midrule
Model Scale & 0.33  &  .57 \\
Korean corpus                                               &  0.30 &  .59 \\
\bottomrule
\end{tabular}
\caption{Results from the ANOVA Test considering two factors. The columns ‘\textit{F}' and ‘\textit{p}' represent the \textit{F-value} and \textit{p-value}, respectively. ‘Korean corpus' refers to the supplemental Korean dataset used during training.}
\label{tab:result_anova} 
}
\end{table}
\paragraph{Model Scale} We use the ANOVA test on \texttt{Polyglot-Ko}, \texttt{KULLM}, \texttt{KoAlpaca}, \texttt{LLaMA} models to assess sensitivity to number of parameters. Test statistics are reported in Table~\ref{tab:result_anova}. 
We find that model scale does not have a statistically significant impact on accuracy ($F_{1,43}=0.33, p=.57$).
\paragraph{Korean Corpus Scale} While the \texttt{KULLM} and \texttt{KoAlpaca} models are fine-tuned with additional Korean corpora based on the \texttt{Polyglot-Ko} models, and the \texttt{LLaMa-Ko} model is a Korean-specialized version of \texttt{LLaMA}, our findings in Table~\ref{tab:result_anova} suggest that additional training on Korean datasets doesn't significantly enhance their comprehension of Korean culture and language($F_{1,54} = 0.30, p=.59$). In fact, the accuracy is even lower than that of their base models. 
It's notable that \texttt{llama-2 chat}, despite deriving only 0.06\% of its training data from Korean sources, outperforms most Korean-specialized models by a small margin.
\\

Our results reveal that despite their extensive pretraining on vast datasets and large scales, \texttt{GPT-3.5-turbo} and \texttt{Claude-2} perform similarly to open-source models in specific categories like \textit{History}, \textit{Economy}, and \textit{Functional knowledge}.
The results and analysis in this section give insights that simply amassing more data and enlarging the model size (currently common practices in LM) may not be the optimal solution for enhancing the cultural intelligence of Language Models in non-English languages.

\section{Discussion}
\subsection{Difficulty Analysis}
\begin{figure}
    \centering
    \includegraphics[width=0.95\columnwidth]{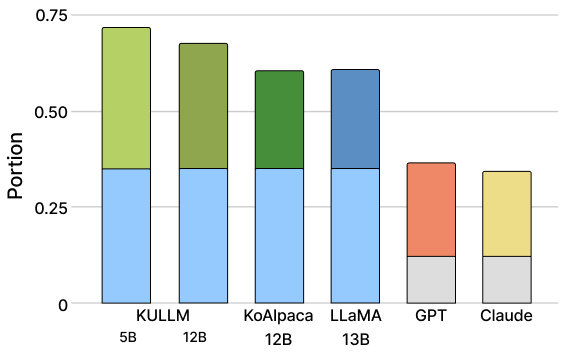}
    \caption{Portions of challenging samples encountered by models of varying sizes. The sky-blue bar represents the shared portion of \texttt{KULLM}, \texttt{KoAlpaca}, and \texttt{LLaMa2-chat}, while the gray bar corresponds to \texttt{GPT-3.5} and \texttt{Claude-2}.}
    \label{fig:Diff_stat}
\end{figure}

\paragraph{Overall Difficulty} \label{overall_diff}
To measure the challenging problems within our dataset, we first define the difficulty by the number of problems with an accuracy below the random selection threshold ($1/N$), where $N$ denotes the number of choices belonging to questions. If the accuracy of the models is below this threshold, we mark it as a challenging problem for a model and vice versa.
As depicted in Figure~\ref{fig:Diff_stat}, open-source models, such as \texttt{Polyglot-Ko}, \texttt{KULLM}, \texttt{KoAlpaca}, \texttt{LLaMa2-Ko}, and \texttt{LLaMa2-chat}, have difficulty with over 60\% of our dataset, with a shared difficulty spanning 35.5\%. LLMs like \texttt{GPT-3.5} and \texttt{Claude-2} face challenges with over 30\%, sharing 12.6\% of problems. Only in 0.6\% of the data were none of the models able to find the correct solution. This emphasizes the CLIcK dataset's inherent complexity.
 ~
\paragraph{Qualitative Study} 
We analyze the tendencies of the models by examining all samples they perform well on and those they don't. We observe no discernible trends that are specific to each category, and observe a wide range of performance across samples derived from the same category and source, even though they have similar levels of difficulty.
For example, Problem 1 and Problem 2 are questions from the Korean society category, derived from KIIP textbook, and both inquire about Korean-specific matters. 
\begin{mdframed}
\scriptsize{\textbf{Problem 1}\\
질문 (Question): \{한국 정부의 주택 정책이 아닌 것은?\\(What is not the Korean government's housing policy?)\}\\
A: \{국민임대주택(National rental housing)\}\\
B: \{공공임대주택(Public rental housing)\}\\
C: \{보금자리주택(Bogeumjari house/Nest house)\}\\
D: \{단기경매주택(Short auction house)\}\\
정답 (Answer): \{단기경매주택(Short auction house)\}
}
\end{mdframed}
\begin{mdframed}
\scriptsize{\textbf{Problem 2}\\
질문 (Question): \{남편이 아내의 오빠를 어떻게 부르는가?\\(How does a husband call his wife's older brother?)\}\\
A: \{처형(Cheohyeong/Sister-in-law)\}\\
B: \{처제(Cheoje/Sister-in-law)\}\\
C: \{처남(Cheonam/Brother-in-law)\}\\
D: \{형님(Hyeongnim/Older brother-in-law)\}\\
정답 (Answer): \{형님(Hyeongnim/Older brother-in-law)\}
}
\end{mdframed}
Problem 2 asks about the everyday life of \textit{Korean society}, whereas problem 1 seeks more professional information, making problem 1 appear more challenging. However, all 13 models achieved 100\% accuracy on Problem 1, yet none correctly answered Problem 2. Therefore, 1) there is less alignment between the model's perceived difficulty and those by humans, and 2) such cultural and linguistic intelligence contexts are challenging for the model to comprehend.
\paragraph{Why Do Models Struggle?}
As mentioned in \S~\ref{prompt}, we conducted a total of \(3N\) experiments for each problem sample, utilizing three different prompts and through cyclic permutation. The model's uncertainty on each sample, calculated using normalized Shannon entropy~\cite{shannon}, is defined as follows:
\begin{equation}
    \text{Uncertainty score} = \text{$-$}\frac{1}{\log N}{\sum_{i \in \text{options}} p_i \log p_i } \nonumber
\end{equation}
\begin{equation}
    \text{where} \quad p_i = \frac{\text{count}(\text{i})}{3N}
\end{equation}
and \(i\) represents each option.

The score is normalized between 0 and 1, considering varying numbers of total options. A score nearer to 0 implies high consistency, whereas one near 1 indicates almost random selection.
For each model's challenging problem, as defined at the beginning of \S~\ref{overall_diff}, we calculate an uncertainty to analyze the reasons behind their low accuracy on our dataset.

\begin{figure}
    \centering
    \includegraphics[width=0.95\columnwidth]{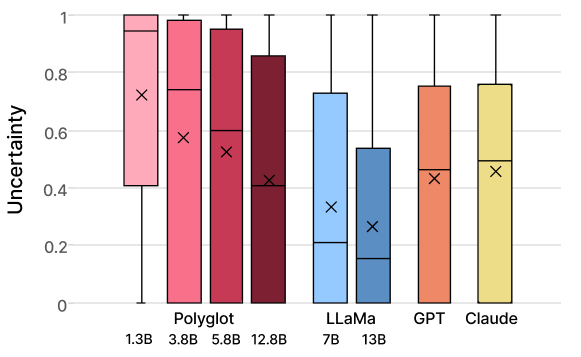}
    \caption{Box-and-whisker plot of uncertainty score of challenging samples across the models. `x' marks the mean, and the horizontal bar represents the median. For \texttt{Polyglot} and \texttt{LLaMa}, consistency rises with increasing model scale. \texttt{Polyglot 1.3B} has a median near 1, while \texttt{LLaMa}'s median is closer to 0.}
    \label{fig:Entropy}
\end{figure}

The plot in Figure~\ref{fig:Entropy} illustrates the uncertainty scores of various models. Smaller models (\texttt{polyglot 1.3B, 3.8B}) tend to randomly select answers without consistency, and as the size increases (12.8B, 5.8B), they tend to choose wrong answers consistently. In other words, models are not getting our dataset wrong because they are unsure of the answer, but they choose specific wrong answers at a high rate. Upon analyzing \texttt{GPT-3.5} and \texttt{Claude-2}, we observe an ambiguous pattern, as their mean and median uncertainty scores both approximate 0.5.

 ~
\subsection{Comparisons to Human Level}
We compare model accuracy to human performance on exams.
Since our dataset does not encompass the actual score distribution for the problems, a simple score conversion,the ratio of correctly answered questions, is applied to facilitate a comparative analysis. We utilize exam statistics from CSAT Korean, TOPIK and Kedu for Korean Culture to assess the models' performances against actual exam takers.
\paragraph{CSAT Korean} An average is taken of the statistics spanning from 2017 to 2020, a total of five years. The CSAT is divided into tiers from 1 (best) to 9 (worst). The \texttt{Polyglot-Ko(12.8B)}, \texttt{KULLM(5.8B, 12.8B)}, and \texttt{KoAlpaca(12B)} models performed at the 9th level, corresponding to the lowest 4\% of Korean high school seniors (3rd-year students). The remaining models are at the 8th grade level, representing the lowest 11\%.
\paragraph{TOPIK} TOPIK is evaluated on an absolute scale, with levels ranging from 1 to 6, with higher being better. \texttt{Claude-2} achieved a level 6, implying it can perform language functions required for specialized research or professional tasks relatively accurately and fluently. While it doesn't reach the proficiency of a native speaker, it doesn't face difficulties in functional performance or expressing meanings. \texttt{GPT-3.5} achieved a level 5, indicating it can appropriately differentiate language use depending on formal and informal, as well as spoken and written contexts. The other models score too low and fall outside the measurement range.
\paragraph{Kedu for Korean Culture} We analyze data over a five-year, from 2014 to 2018. All model results fall below the average of participants, which stood at 49.9\% correct answers. \texttt{Claude-2} and \texttt{GPT-3.5} are lower than the average by about 10\%, scoring 39.1\% and 37.0\%, respectively. Meanwhile, other models lagged by more than 20\% compared to the average.
\section{Conclusion}
We introduced CLIcK, A Benchmark Dataset of Cultural and Linguistic Intelligence in Korean. CLIcK emerges as a uniquely Korean-centric dataset sourced from Korean examinations and textbooks. The dataset is categorized into two main category and 11 sub-categories, which enable fine-grained and Korean-centric evaluation. 
Through our analyses and experiments, we observed that five open-source models struggle with over 60\% of the data. Proprietary LLMs outperform these models yet still require further improvement. Interestingly, simply scaling up the model or fine-tuning it with additional Korean corpora doesn't guarantee enhanced Korean linguistic and cultural knowledge of models. This implies that models find it challenging to understand non-English linguistic and cultural intelligence, highlighting the need for more tailored methods in further research.

\section*{Ethics Statement}

This work presents CLIcK dataset, a free and open evaluation benchmark of cultural and linguistic intelligence in Korean. Our dataset contains multiple-choice QAs with four or five choices.
The data sources are publicly available exams and published articles from the Korean government, and we receive official permission from the related agencies. We scrutinize all samples in the dataset to ensure that no personally identifiable information or sensitive content is included.
Furthermore, we have made our best effort to deliver Korean culture correctly by 1) generating samples from reliable grounds and 2) iteratively validating samples.

\section*{Acknowledgements}
This project was funded by Institute of Information communications Technology Planning Evaluation (IITP) grant funded by the Korea government(MSIT) (No. 2022-0-00184, Development and Study of AI Technologies to Inexpensively Conform to Evolving Policy on Ethics).

\section*{Bibliographical References}\label{sec:reference}

\bibliographystyle{lrec-coling2024-natbib}
\bibliography{references/anthology, references/custom}

\section*{Language Resource References}
\bibliographystylelanguageresource{lrec-coling2024-natbib}
\bibliographylanguageresource{references/anthology, references/languageresource}

\clearpage

\newpage
\section*{Appendix}
\appendix
\section{Exam Description}
\label{sec:appendix_A}
\paragraph{College Scholastic Ability Test of Korea (CSAT)}\thinspace\footnote{\url{https://www.suneung.re.kr/}}
CSAT, endorsed by Korean universities, assesses scholastic aptitude based on the Korean high school curriculum and is administered annually by the Korean Ministry of Education.


\paragraph{Test of Proficiency in Korean (TOPIK)}\thinspace\footnote{\url{https://www.topik.go.kr/}}

TOPIK measures and evaluates the proficiency of learners of Korean as a second language.


\paragraph{National Public Service Examination - Grade 9 (PSE)}\thinspace\footnote{\url{https://www.gosi.kr/}\label{gosi}}
PSE evaluates the knowledge and abilities of individuals who wish to work in the public sector. 
We use the history section and the language section which assesses proficiency in grammar, vocabulary, and reading comprehension.


\paragraph{Public Service Aptitude Test (PSAT)}\thinspace\textsuperscript{\ref{gosi}}
PSAT assesses the aptitudes essential for performing public duties at a higher standard than the PSE.
Our study focuses on the Korean Constitution section of the PSAT subjects.


\paragraph{Korean History Exam (KHE)}\thinspace\footnote{\url{https://www.historyexam.go.kr}}
KHE measures the historical literacy of Korean citizens.


\paragraph{Test of Teaching Korean as a Foreign Language (Kedu)} \thinspace\footnote{\url{https://www.q-net.or.kr}}
Kedu certifies individuals aspiring to teach Korean to overseas Koreans or foreigners. It covers both Korean language and culture.




\paragraph{The Korean Immigration and Integration Program (KIIP)} \footnote{\url{https://www.immigration.go.kr}} KIIP assists foreigners in integrating into Korean society. We use the basic level KIIP textbook and generate QA pairs using the contents. 


\section{Question Generation Using GPT-4}
\label{sec:appendix_B}
We employed the GPT-4 language model for the generation of multiple-choice questions, focusing on content extracted from the Korean Integration and Immigration Program (KIIP) textbooks. The process involved several key steps to ensure the generation of high-quality, relevant educational questions.

\paragraph{Text extraction}
We extracted text from the KIIP textbooks to serve as the basis for our question generation. To manage this extensive textual data effectively, the extracted text was split into smaller, manageable chunks. This splitting was accomplished using a \textit{RecursiveCharacterTextSplitter} from Langchain. The parameters for this splitter were chosen based on initial experiments to balance between maintaining textual coherence and ensuring manageable chunk sizes for processing.
\paragraph{Prompt}
The core of our question generation process involved prompting GPT-4 with a specific structure to ensure that the questions generated were diverse, relevant, and adhered to a specific format. The prompt used for GPT-4 was as follows:

\begin{table}[h]
\resizebox{\columnwidth}{!}{
\begin{tabular}{l}
\hline
Original Korean Prompt\\ \hline

\begin{tabular}[c]{@{}l@{}}다음 제시문을 읽고 4지 선다 문제 10개를 만들어줘.\\문제를 만들 때 내용은 겹치지 않게 하고 형식은 다음을 포함한 \\json 형식으로 만들어줘. question\_id는 \{current\_cnt\}부터 시작해서 \\1부터 증가해줘.\\ "cite": 제시문에서 문제를 만드는 데 사용한 문장\\ "question\_id": \{문제 번호\}\\ "question": \{문제\}\\ "choices": \{보기\}\\ "answer": \{답\}\\ "제시문": \{content\}\end{tabular} \\ \hline
\end{tabular}
}
\end{table}

\begin{table}[h]
\resizebox{\columnwidth}{!}{
\begin{tabular}{l} 
\hline
English Translated Prompt \\ \hline

\begin{tabular}[c]{@{}l@{}}Read the following passage and create 10 multiple-choice \\questions based on it.\\ Ensure that the content of each question does not overlap and format\\ them in JSON format including the following elements.\\ The question\_id should start from \{current\_cnt\} and increase by 1 there-\\after.\\
"cite": The sentence from the passage used to create the question\\
"question\_id": \{Question number\}\\
"question": \{Question\} \\
"choices": \{Options\} \\
"answer": \{Answer\} \\
"Passage":
\{content\}\end{tabular} \\ \hline
\end{tabular}
}
\end{table}

This prompt was designed to instruct GPT-4 to produce a set of 10 multiple-choice questions for each text chunk, each with a unique question identifier ("question\_id"). The format of the prompt ensured that the questions did not overlap in content and were presented in a structured JSON format. This format included a citation from the text ("cite"), the question itself ("question"), multiple choices ("choices"), and the correct answer ("answer"). This citation was not just for referencing purposes but also served a vital role in the validation process. It allowed human reviewers to quickly ascertain the validity of the generated question and to confirm that the provided answer was indeed correct according to the textbook content.

\paragraph{Verification}As a final step in the process, any instances generated by GPT-4 that did not conform to the specified format were identified and removed. 


\end{document}